%% file: main.tex
\documentclass[10pt,twocolumn,letterpaper]{article}

\usepackage[pagenumbers]{cvpr}      

\input{preamble}

\definecolor{cvprblue}{rgb}{0.21,0.49,0.74}
\usepackage[pagebackref,breaklinks,colorlinks,citecolor=cvprblue]{hyperref}

\def\paperID{10800} 
\def\confName{CVPR}
\def\confYear{2024}

\title{Towards Faster Training of Diffusion Models: An Inspiration of A Consistency Phenomenon}

\author{Tianshuo Xu\\
HKUST(GZ)\\
{\tt\small txu647@connect.}\\{\tt\small hkust-gz.edu.cn}
\and
Peng Mi\\
The University of Sydney\\
{\tt\small mipeng01@gmail.com}
\and
Ruilin Wang\\
Xiamen University\\
{\tt\small 31520231154317}\\{\tt\small@stu.xmu.edu.cn}
\and
Yingcong Chen\\
HKUST(GZ)\\
{\tt\small yingcongchen@ust.hk}
}

\begin{document}
\maketitle
\input{sec/0_abstract}    
\input{sec/1_intro}
\newpage
{
    \small
    \bibliographystyle{ieeenat_fullname}
    \bibliography{main}
}

\input{sec/X_suppl}

\end{document}

%% file: preamble.tex
\usepackage[dvipsnames]{xcolor}


%% file: sec/0_abstract.tex
\begin{abstract}
Diffusion models (DMs) are a powerful generative framework that have attracted significant attention in recent years. However, the high computational cost of training DMs limits their practical applications. In this paper, we start with a consistency phenomenon of DMs: we observe that DMs with different initializations or even different architectures can produce very similar outputs given the same noise inputs, which is rare in other generative models. We attribute this phenomenon to two factors: (1) the learning difficulty of DMs is lower when the noise-prediction diffusion model approaches the upper bound of the timestep (the input becomes pure noise), where the structural information of the output is usually generated; and (2) the loss landscape of DMs is highly smooth, which implies that the model tends to converge to similar local minima and exhibit similar behavior patterns. This finding not only reveals the stability of DMs, but also inspires us to devise two strategies to accelerate the training of DMs. First, we propose a curriculum learning based timestep schedule, which leverages the noise rate as an explicit indicator of the learning difficulty and gradually reduces the training frequency of easier timesteps, thus improving the training efficiency. Second, we propose a momentum decay strategy, which reduces the momentum coefficient during the optimization process, as the large momentum may hinder the convergence speed and cause oscillations due to the smoothness of the loss landscape. We demonstrate the effectiveness of our proposed strategies on various models and show that they can significantly reduce the training time and improve the quality of the generated images.
\end{abstract}

%% file: sec/1_intro.tex
\section{Introduction}
Diffusion Models (DMs) \cite{sohl2015deep, ho2020denoising, song2020score, song2020denoising}, a prominent class of generative models, have garnered considerable attention in recent years. Owing to their exceptional capability to model intricate data distributions, DMs have catalyzed significant advancements in various domains. These include image generation \cite{nichol2021improved, dhariwal2021beatgans, rombach2021highresolution}, inpainting \cite{lugmayr2022repaint, anciukevivcius2023renderdiffusion}, image manipulation \cite{zhang2023controlnet, lugmayr2022repaint, kawar2023imagic}, video generation \cite{ho2022imagen, blattmann2023align, wang2023videocomposer}, and speech synthesis \cite{jeong2021diff_speech, zhang2023survey_speech}. 

However, DMs are constrained by their high computational demands, which hamper their practical applications. For example, Guided-Diffusion \cite{dhariwal2021beatgans} requires 1878 V100-hours to train a model at 512 resolution, whereas BigGAN-deep \cite{Brock2018LargeSG_biggan} only needs 256-512 V100-hours for the same resolution, which is four to eight times less than DMs. Although some recent methods, such as ControlNet \cite{zhang2023controlnet} and LoRA \cite{hu2022lora}, can leverage pre-trained large models to efficiently fine-tune them for downstream tasks, these methods are still not applicable to some complex scenarios.

\begin{figure*}[t]
  \centering
  \includegraphics[width=0.9\linewidth]{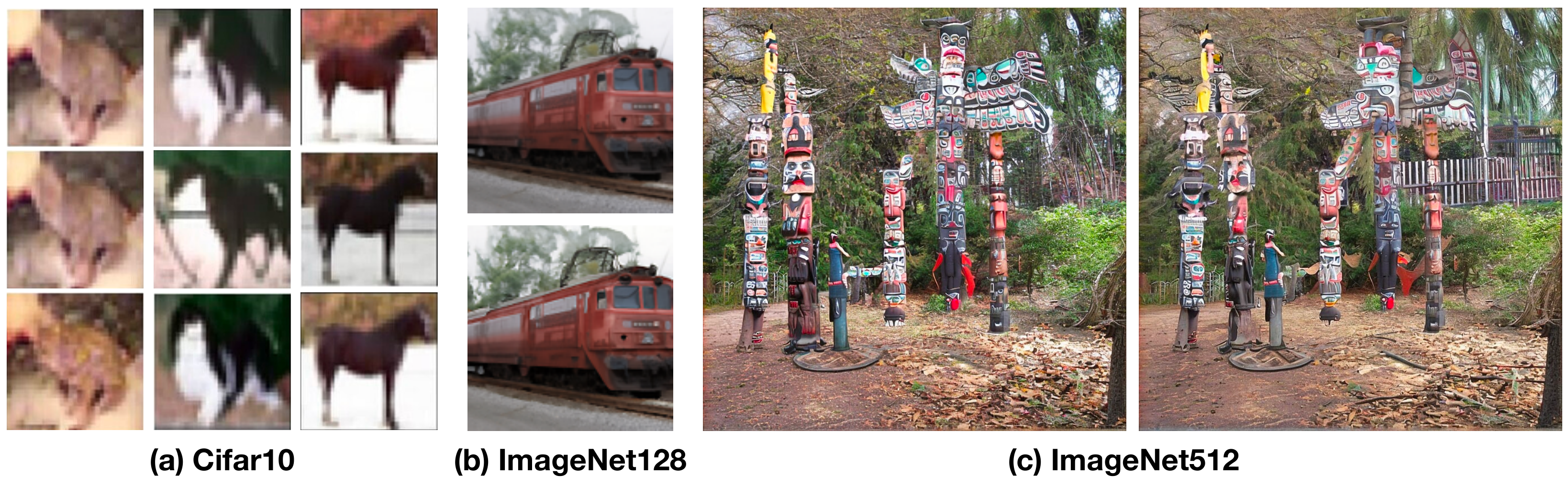}
  \caption{Illustration of the consistency phenomenon in diffusion models (DMs). Despite different initializations or structural variations, DMs trained on the same dataset produce remarkably consistent results when exposed to identical noise during sampling. (a) presents three models \cite{nichol2021improved} trained on CIFAR10 with different initializations. (b) depicts two models \cite{dhariwal2021beatgans} trained on ImageNet128 with different structures. (c) showcases the large and huge models of UViT \cite{bao2023uvit} trained on ImageNet512.}
  \label{fig:consistency}
\end{figure*}

In this paper, we propose to accelerate the convergence speed of diffusion models by exploiting some properties that we discovered in a consistency phenomenon. The consistency phenomenon of DMs, as illustrated in Fig.~\ref{fig:consistency}, reveals that trained DMs with different initialization or even different architecture can generate remarkably similar outputs given the same noise inputs, which is a rare property in other generative models. We then attribute this phenomenon to two factors: (1) the learning difficulty of DMs is lower when the noise-prediction diffusion model reaches the upper limit of the time step (the input becomes pure noise), where the output tends to form the structural information; and (2) the loss landscape of DMs is extremely smooth, which implies that the model is likely to converge to similar local minima and exhibit similar behavior patterns. This finding not only unveils the stability of DMs, \emph{i.e.,} DMs are stable in learning the noise-to-data mapping, but also inspires us to design two strategies to speed up the training of diffusion models.



Since the learning difficulty of DMs can be explicitly indicated by the noise ratio, that is, for noise-prediction DMs, the higher the noise, the easier to learn, which aligns well with the principle of curriculum learning that advocates learning from easy to hard, we propose a curriculum learning based time step schedule (CLTS), which aims to gradually decrease the sampling probabilities of easy-to-learn timesteps and increase the probabilities of important ones, thus improve training efficiency. 


We also propose a momentum decay with learning rate compensation (MDLRC), which exploits the high smoothness of the loss landscape of DMs. Unlike GANs \cite{Brock2018LargeSG_biggan}, which require a large momentum to ensure gradient stability, DMs can benefit from a smaller momentum. Our experimental results show that a large momentum may hinder the convergence speed and cause oscillations of DMs. Therefore, decreasing the momentum can further improve the convergence efficiency.



We evaluate our optimization methods on various DMs, such as Improved-Diffusion \cite{nichol2021improved} and Guided-Diffusion \cite{dhariwal2021beatgans}, and demonstrate that they can enhance the convergence speed of DMs. For instance, on ImageNet128 \cite{deng2009imagenet}, our methods achieve a 2.6$\times$ speedup in training with standard Guided Diffusion \cite{dhariwal2021beatgans} and a 2$\times$ speedup on CIFAR10 \cite{krizhevsky2009cifar} compared with standard Improved Diffusion \cite{nichol2021improved}, showing the effectiveness of our methods. 

In summary, our main contributions are three-fold:
\begin{itemize}
  \item Discovering and explaining the consistency phenomenon of DMs, which shows that DMs with different initializations or architectures can produce similar outputs given the same noise inputs.
  \item Proposing two strategies to accelerate the training of DMs based on the properties found in consistency phenomenon: a curriculum learning based timestep schedule and a momentum decay strategy.
  \item Evaluating the effectiveness of the proposed strategies on various models and datasets, and showing that they can significantly reduce the training time and improve the quality of the generated images.
\end{itemize}

\begin{figure*}[t]
  \centering
  \includegraphics[width=0.9\linewidth]{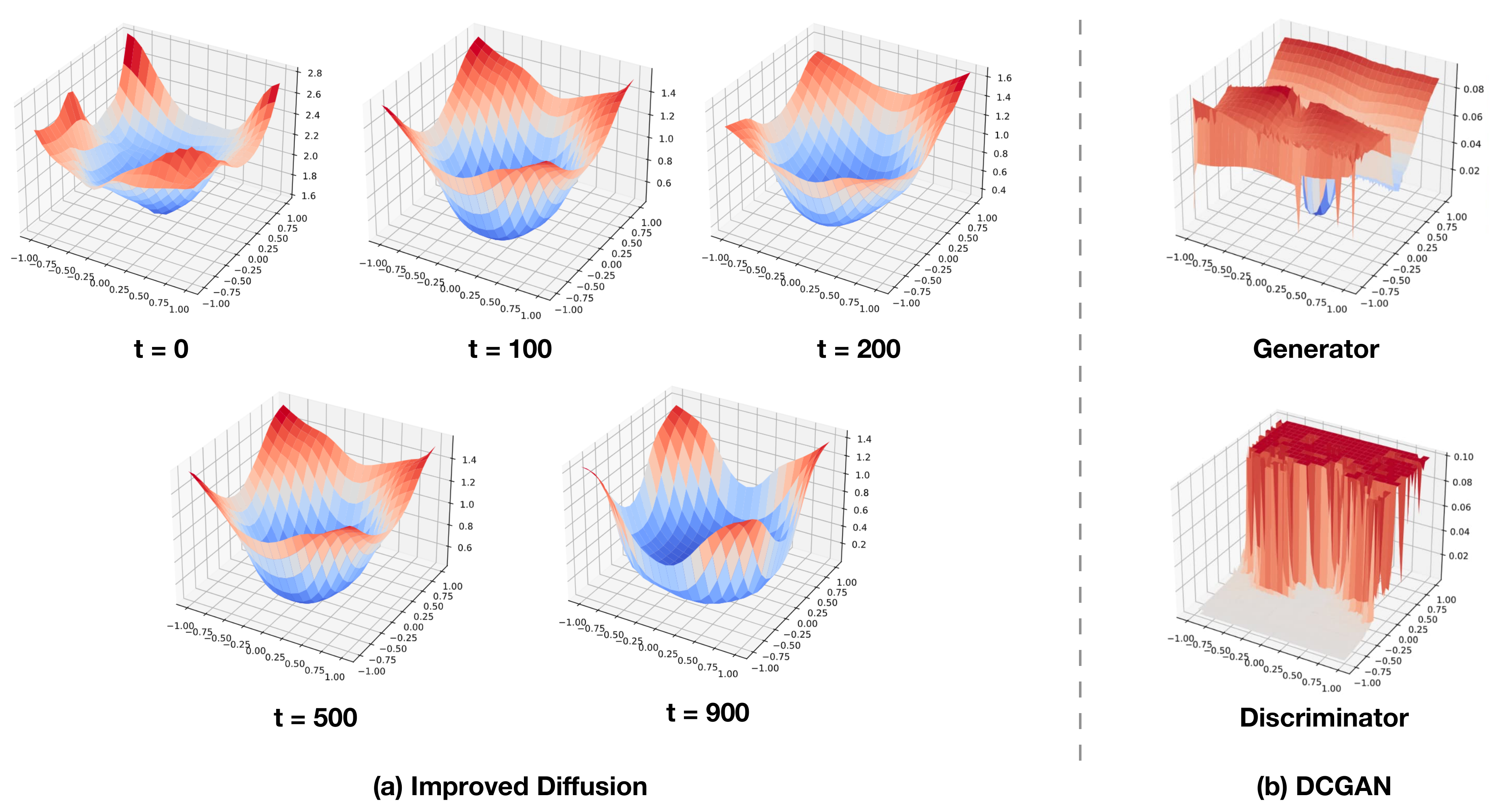}
  \caption{Visualization of the loss landscapes of Improved Diffusion \cite{nichol2021improved} and DCGAN \cite{radford2015dcgan}, where \textbf{t} is the timestep of DMs. Both models were trained on the CIFAR10 dataset \cite{krizhevsky2009cifar}. Obviously, the loss landscape of DMs is smoother compared to GANs. More landscapes of DM and GAN can be viewed at Appendix~\ref{app:loss_land}.}
  \label{fig:loss_land}
\end{figure*}

\section{Related Work}
\subsection{Diffusion Models}

Diffusion Models (DMs) are a class of generative models that use techniques from non-equilibrium thermodynamics to learn the latent structure of complex data distributions. They were first introduced by \cite{sohl2015deep}, who applied their method to image and text generation. Later, \cite{ho2020denoising} proposed Denoising Diffusion Probabilistic Models (DDPM), which improved the sampling efficiency and quality of diffusion models by using a denoising score matching objective and a learned diffusion process. \cite{rombach2021highresolution} developed Latent Diffusion Models (LDMs), which compressed high-resolution images into lower-dimensional representations using pretrained autoencoders. They also introduced cross-attention layers into the model architecture, which enabled LDMs to handle various conditioning inputs, such as text or bounding boxes, and generate high-resolution images in a convolutional manner. Despite the success of DMs using UNet \cite{ho2020denoising, nichol2021improved, rombach2021highresolution}, a convolutional neural network, \cite{bao2023uvit} and \cite{Peebles2022ScalableDM_dit} discovered the feasibility of using Vision Transformer \cite{Dosovitskiy2020ViT} in DMs, achieving state-of-the-art generation results.


Several studies have focused on improving the DMs from various aspects \cite{Karras2022ElucidatingTD, Chen2023OnTI}. \cite{nichol2021improved} proposed several techniques to enhance the performance and efficiency of DMs, such as employing a learned variance schedule, adopting a cosine timestep schedule for low-resolution data, and developing a multi-scale architecture. \cite{dhariwal2021beatgans} further improved the performance and fidelity of DMs, by incorporating advanced design concepts of BigGAN \cite{Brock2018LargeSG_biggan}. Although integrating the sophisticated model structure of GAN can benefit the performance of DMs, they also adopt the same large momentum setting, which is sub-optimal, because the loss landscape of DMs is highly smoothed. A large momentum not only affects convergence efficiency but also causes oscillations. We discussed this in detail in Section \ref{subsec:loss_land}.

\subsection{Training Acceleration}
Learning from easy to hard is the core concept of curriculum learning~\cite{Bengio2009CurriculumL},~\emph{e.g.}, Curriculum Learning improves language translation model performance by up to 2.2 and reduces training time by up to 70\%~\cite{nlpcurriculum}. Soviany \emph{et al.} \cite{soviany2020image} train a support vector regression model to assess the difficulty of images, and subsequently, training with the selected images results in improved stability and superior performance. Ghasedi \emph{et al.} \cite{ghasedi2019balanced} proposes a heuristic strategy to assign weights to individual images during the training process.

Additionally, sparse training is an effective technique for enhancing network performance and accelerating training speed~\cite{mi2023systematic}. Mi \emph{et al.} \cite{mi2023systematic} applies a series of hardware-friendly sparse masks to the variables during the forward and reverse processes to achieve acceleration. Another option is to consider acceleration from the perspective of an optimizer~\cite{Shampoo, kfac, mfac, liu2023sophia}. Frantar~\emph{et al.}~\cite{mfac} presents a highly efficient algorithm for approximating matrices, specifically designed for utilization in second-order optimization.

Momentum decay is another promising technique for accelerating the training process. Chen \emph{et al.} \cite{Chen2019DecayingMH} introduced Decaying Momentum (Demon), a method that reduces the cumulative effect of a gradient on all future updates. Demon outperformed other methods in 28 different settings, involving various models, epochs, datasets, and optimizers. This work inspired our proposed optimization method.

Recently, motivated by the large computation cost in training DMs, several approaches focus on training acceleration of DMs \cite{Hang2023EfficientDT, Wu2023FastDM, Choi2022PerceptionPT}. Hang \emph{et al.} \cite{Hang2023EfficientDT} treat the diffusion training as a multi-task learning problem, and introduce a simple yet effective approach called Min-SNR, which adapts the loss weights of timesteps based on clamped signal-to-noise ratios. Wu \emph{et al.} \cite{Wu2023FastDM} proposed the momentum-based diffusion process, which can be modeled as a damped oscillation system, whose critically damped state has the optimal noise perturbation kernel that avoids oscillation and accelerates the convergence speed.

\begin{table*}[tb]
\centering
\caption{Comparing consistencies of DMs and GANs in learning noise-to-data mapping.}
\label{tab:consistency}
\scalebox{0.95}{
\begin{tabular}{ccccc}
\hline
& Datasets    & \begin{tabular}[c]{@{}c@{}}Different \\ Initializations\end{tabular} & \begin{tabular}[c]{@{}c@{}}Different \\ Structures\end{tabular} & Consistency (PSNR) \\
\hline 
Improved Diffusion & Cifar10     &       \checkmark     &                      &     \textbf{20.14}    \\
DCGAN              & Cifar10     &        \checkmark     &                      &    10.48     \\
Guided Diffusion   & ImageNet128 &       \checkmark      &     \checkmark     &       \textbf{17.23}   \\
BigGAN             & ImageNet128 &       \checkmark         &   \checkmark    &     8.58      \\
U-ViT              & ImageNet512 &       \checkmark       &     \checkmark    &     \textbf{14.37}     \\
BigGAN             & ImageNet512 &      \checkmark     &      \checkmark    &       6.40   \\
\hline
\end{tabular}}
\end{table*}

\section{The Consistency Phenomenon of Diffusion Models}




The consistency phenomenon refers to the observation that trained DMs with different initializations or even different architectures can generate remarkably similar outputs given the same noise inputs.
In this section, we investigate the consistency phenomenon of DMs. We begin by quantifying the strength of this phenomenon in DMs and GANs in Sec.~\ref{subsec:mapping}, and show that it is rare in GANs. Next, we uncover the underlying mechanism of the consistency phenomenon in DMs through a formulation analysis in Sec.~\ref{subsec:uncover}, and reveal that it is related to the noise-prediction capability of DMs. Finally, we demonstrate that the loss landscape of DMs is highly smooth in Sec.~\ref{subsec:loss_land}, which implies that DMs are prone to converge to similar local minima and exhibit consistent behavior patterns.


\subsection{Quantifying the Consistency Phenomenon}\label{subsec:mapping}

In this section, we investigate the consistency phenomenon of DMs by conducting a quantification experiment. We choose three diffusion frameworks as representative DMs: Improved Diffusion \cite{nichol2021improved}, Guided Diffusion \cite{dhariwal2021beatgans}, and U-ViT \cite{bao2023uvit}. We also select two GAN frameworks: DCGAN \cite{radford2015dcgan} and BigGAN \cite{Brock2018LargeSG_biggan} for comparison. We train DMs and GANs with different hyper-parameters on three benchmarks: Cifar10 \cite{krizhevsky2009cifar}, ImageNet128 \cite{deng2009imagenet}, and ImageNet512 \cite{deng2009imagenet}. For BigGAN, we use BigGAN (128/512) and BigGAN-deep (128/512) models. We present the results of the consistency experiment and the detailed settings in Table~\ref{tab:consistency}. We show the generated images in Appendix~\ref{app:illus_consistency}.


In the quantification experiment, we use the peak signal-to-noise ratio (PSNR) to measure the consistency of each model. Specifically, for a group of models, \emph{e.g.,} different initializations of Improved Diffusion or large and huge models of U-ViT, we sample 32 images with the same sampling seed, which means the initial noise and the noise per round are the same. Suppose we have $N$ models of a group, each model generate $M$ images, we then measure the consistency $C(\cdot)$,
\begin{equation}
    C(q) = \frac{1}{M}\sum_{i=1}^{M} \frac{1}{N-1}\sum_{j=2}^N\ \text{PSNR}(q_{i,1},\ q_{i,j}),
\end{equation}
where $q \in \mathbb{R}^{N\times M}$ is the matrix of images.

The result of the consistency experiment reveals that all DMs have much higher consistency than GANs, regardless of the dataset, initialization, or model structure.


\subsection{Uncovering the Consistency Phenomenon}\label{subsec:uncover}
From the formula of DMs, it is not difficult to find that the model is easy to trivial when $t \to T$, we start with the formulation of the forward diffusion process, derived from Eq.~\ref{eq:forward_single} (Appendix~\ref{app:preliminaries}),
\begin{equation}
    x_t = \sqrt{\bar{\alpha}_t}\cdot x_0 + \sqrt{1 - \bar{\alpha}_t}\cdot \epsilon,\  \epsilon \in \mathcal{N}(0, \textbf{I}),
\end{equation}
where $\alpha_t = 1 - \beta_t$ and $\bar{\alpha}_t = \Pi_{s=0}^t \alpha_s$. 
Note that, $\bar{\alpha}_t$ is a factor, ranging from 0 to 1, when $t\to T$, $\bar{\alpha}_t \to 0$ then $x_t \to \epsilon$. According to Eq.~\ref{eq:simple_loss} (Appendix~\ref{app:preliminaries}), 
\begin{equation}
    L_{\text{simple}} = E_{x_0\sim q(x_0), \epsilon\sim \mathcal{N}(0, \textbf{I})}\left[||\epsilon - \epsilon_\theta(x_t, t)||^2 \right],
\end{equation}
when $x_t \to \epsilon$, 
\begin{equation}
    \epsilon_\theta \to \textbf{I},
\end{equation}
which means the $\epsilon$-predicted DMs tend to be trivial when $t \to T$. 

The observation of the results of the consistency experiments supports the above derivation. We find that the images (Fig.~\ref{fig:consistency}) with consistency phenomenon are highly similar in structure but different in detail, while the timesteps when the diffusion model generates structural information are $t \to T$ (Fig.~\ref{fig:d_process}, in Appendix). 

To further validate the cause of the consistency phenomenon is the $\epsilon$-predicted mechanism leading to model triviality when $t \to T$. We train an $x_0$-predicted DM as a counterexample, we change the loss function as
\begin{equation}
    L_{x_0} = E_{x_0\sim q(x_0)}\left[||x_0 - \mu_\theta(x_t, t)||^2 \right].
\end{equation}
As shown in Fig.~\ref{fig:x0} (in Appendix), the consistency phenomenon disappears. Thus, we conclude that the cause of the consistency phenomenon is the $\epsilon$-predicted mechanism leading to model triviality when $t \to T$.

\subsection{Analyzing the Smoothness of Loss Landscape}\label{subsec:loss_land}

The previous conclusion only demonstrates that the consistency phenomenon is caused by the noise-prediction mechanism of DMs, which have lower learning difficulty when $t \to T$. However, this does not explain why DMs behave similarly when $t \to T$. Therefore, in this subsection, we aim to investigate the loss landscape of the DMs. By comparing with GANs, we observe that the loss landscape of the DMs is highly smooth, so the DMs are more likely to converge to similar local minima, and thus exhibit similar behavior patterns.

Note that, due to the high dimensionality of the models’ parameters, it is impractical to access the full information of the loss landscape. Therefore, we adopt a partial analysis based on 1D interpolation of models and hessian spectra, following the method proposed by \cite{li2018visualizing}.


\begin{figure*}[tbh]
    \centering
    \includegraphics[width=0.8\linewidth]{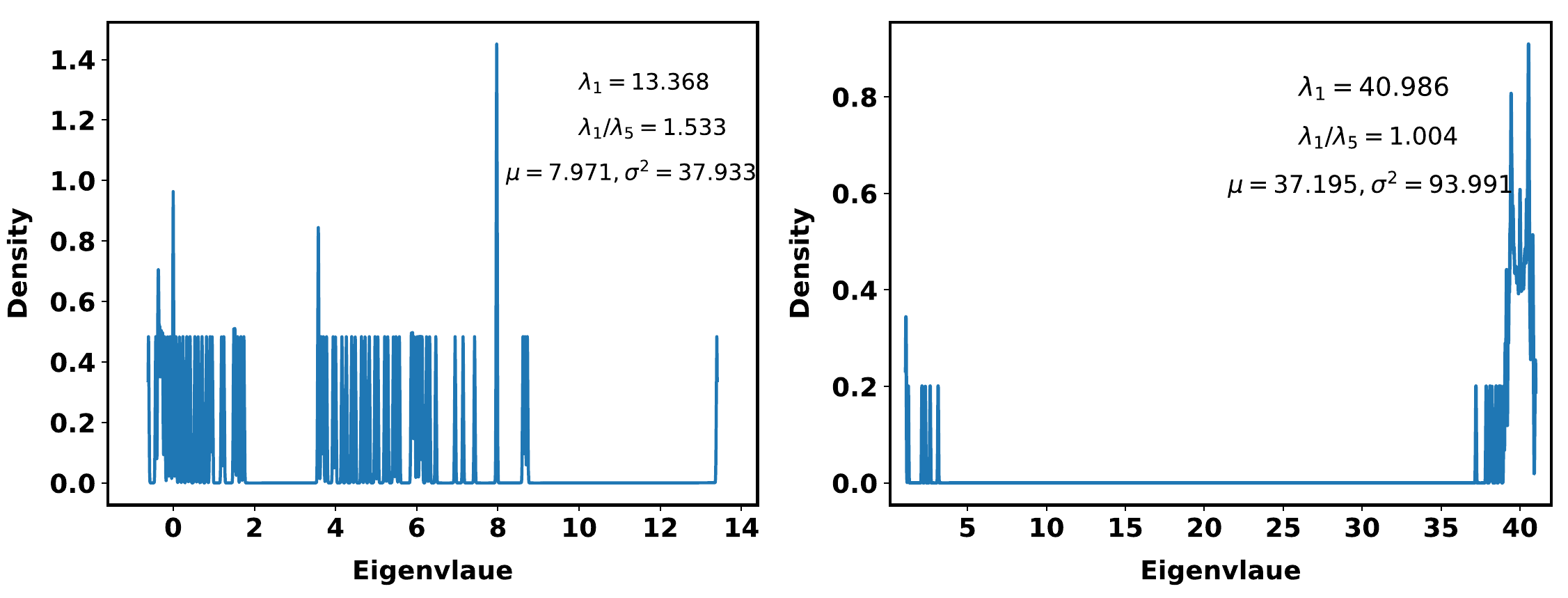}
    \caption{Illustration of the Hessian spectrum of DMs(left) and GANs(right). $\lambda_i$ is the $i$-th largest eigenvalue and $\mu$ and $\sigma$ is the mean and variance of eigenvalue respectively. The larger the dominant eigenvalue, the sharper the landscape, and the greater the differences among eigenvalues, the more difficult the model is to optimize.}
    \label{fig:hessian}
\end{figure*}

\begin{figure}[htb]
    \centering
    \includegraphics[width=\linewidth]{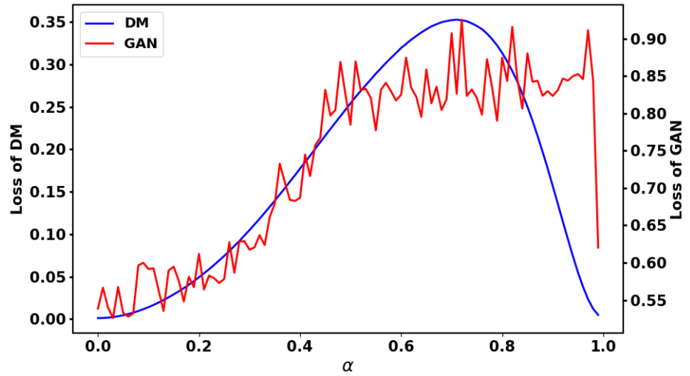}
    \caption{Illustration of the 1D-interpolation results of DMs and GANs. The jitter red line indicates the geometry of GAN's landscape is rougher.}
    \label{fig:1d_inter}
\end{figure}

\textbf{1D interpolation} is a technique that generates new data points by leveraging existing data. In our research, we employed 1D linear interpolation to estimate the position $\boldsymbol{\theta}$ within the landscape using the provided models at different stages of training, namely $\boldsymbol{\theta}_a$ and $\boldsymbol{\theta}_b$. This involved calculating the weighted sum of these two models.
\begin{align}
    \boldsymbol{\theta} = \alpha \boldsymbol{\theta}_a + (1 - \alpha) \boldsymbol{\theta}_b. \quad (0 \leq \alpha \leq 1)
\end{align}
We use interpolation to analyze the relationship between different training stages and gather valuable information. Our approach involves training a diffusion model and a GAN model, followed by selecting models from various training steps as anchor points. Specifically, we select the models trained 10 and 100 epochs for both DM and GAN. These selections, shown in Fig.~\ref{fig:1d_inter}, represent models from both early and late convergence stages.

As shown in Fig.~\ref{fig:1d_inter}, the GAN model exhibits more erratic changes in loss, indicating that changes in GAN parameters lead to relatively larger changes. 


\textbf{Hessian spectra} refers to the distribution of eigenvalues in the Hessian matrix. Inspired by the connection between the geometry of the loss landscape and the eigenvalue, we approximate the Hessian spectrum by the Lanczos algorithm and the results of Diffusion and GAN are shown in Fig.~\ref{fig:hessian}. From the figure, it can be seen that the dominant eigenvalue of GAN's is larger,~\emph{i.e.}, $\lambda_1=13.3$(DMs) \emph{v.s.} $\lambda_1=40.9$(GANs), and dispersion,~\emph{i.e.}, $\sigma^2=37.9$(DMs) \emph{v.s.} $\sigma^2=93.9$(GANs), which implies that the landscape of GAN is steeper and more rugged, which also means that the GAN is more difficult to optimize.


\section{Optimization}

In this section, we discuss how we can exploit the properties above to optimize the DMs. Inspired by curriculum learning \cite{Bengio2009CurriculumL}, we propose a timestep schedule that gradually decreases the sampling probabilities of timesteps $t \to T$ as the training progresses (in Section~\ref{subsec:cl}). Next, in Section~\ref{subsec:md}, we provide details of our optimal momentum schedule based on the highly smoothed loss landscape of diffusion models.

\subsection{Optimization of Timestep Schedule}\label{subsec:cl}
The consistency phenomenon indicates that the learning difficulty of DMs has an explicit relation with the noise rate, which means when timesteps $t \to T$, DMs are easy to converge. Therefore, we propose a novel approach to improve the training efficiency of DMs. We observe that the mainstream diffusion frameworks treat every timestep equally and use uniform probability $U(t) = 1/T$ to sample timesteps during the training, which leads to redundant training for $t \to T$.

To address this issue, we adopt curriculum learning \cite{Bengio2009CurriculumL}, a training acceleration technique that is based on the principle of learning from easy to hard. Coincidentally, DMs have the natural ability to produce data with varying levels of difficulty. 

We propose the curriculum learning based timestep schedule (CLTS), which aims to gradually decrease the sampling probabilities of timesteps $t \to T$ as the training progresses and increase the probabilities of others, \emph{i.e.,} find an optimal timestep distribution.
For simplicity, we assume that the optimal timestep distribution follows a Gaussian distribution, 
\begin{equation}\label{eq:gaussian}
    N(t) = \frac{1}{\sigma \sqrt{2\pi}}\exp \left(-\frac{(t-\mu)^2}{2\sigma^2}\right),
\end{equation} 
where the mean $\mu$ indicates the most important timesteps and others are less important. To further reduce the number of hyper-parameters in CLTS, we set the variance $\sigma = T$, which means the distribution has a standard variance across timesteps.

Our initial idea was to shift the Gaussian distribution as the mean moves from $T$ to $0$, but this achieved little improvement. We conjecture that this is because the generation of the DMs requires the involvement of all the timesteps. The strategy of shifting the distribution leads to the sampling probabilities of all timesteps except for $t \to T$ being too small at the initial stage. Thus, we propose a mixed distribution, introducing a factor $\gamma$ to transfer the distribution from uniform to Gaussian,
\begin{equation}\label{eq:gamma}
P(t) = (1-\gamma) U(t) + \gamma N(t),\ \gamma=\frac{\text{current iteration}}{\text{target iteration}},
\end{equation}
where the target iteration is a hyper-parameter that adjusts the speed of the Gaussian distribution emerging.

It is worth noting that our proposed CLTS has a similar implementation to \cite{Hang2023EfficientDT} and \cite{Choi2022PerceptionPT}. However, there are significant differences in our underlying philosophies. Inspired by curriculum learning, our method is based on the principle of learning from easy to hard, while \cite{Hang2023EfficientDT} and \cite{Choi2022PerceptionPT} are focused on finding an optimal distribution. Our method is much more robust and efficient in extensive experiments (which we show in the next section).

\begin{figure*}[thb]
  \centering
  \includegraphics[width=0.85\linewidth]{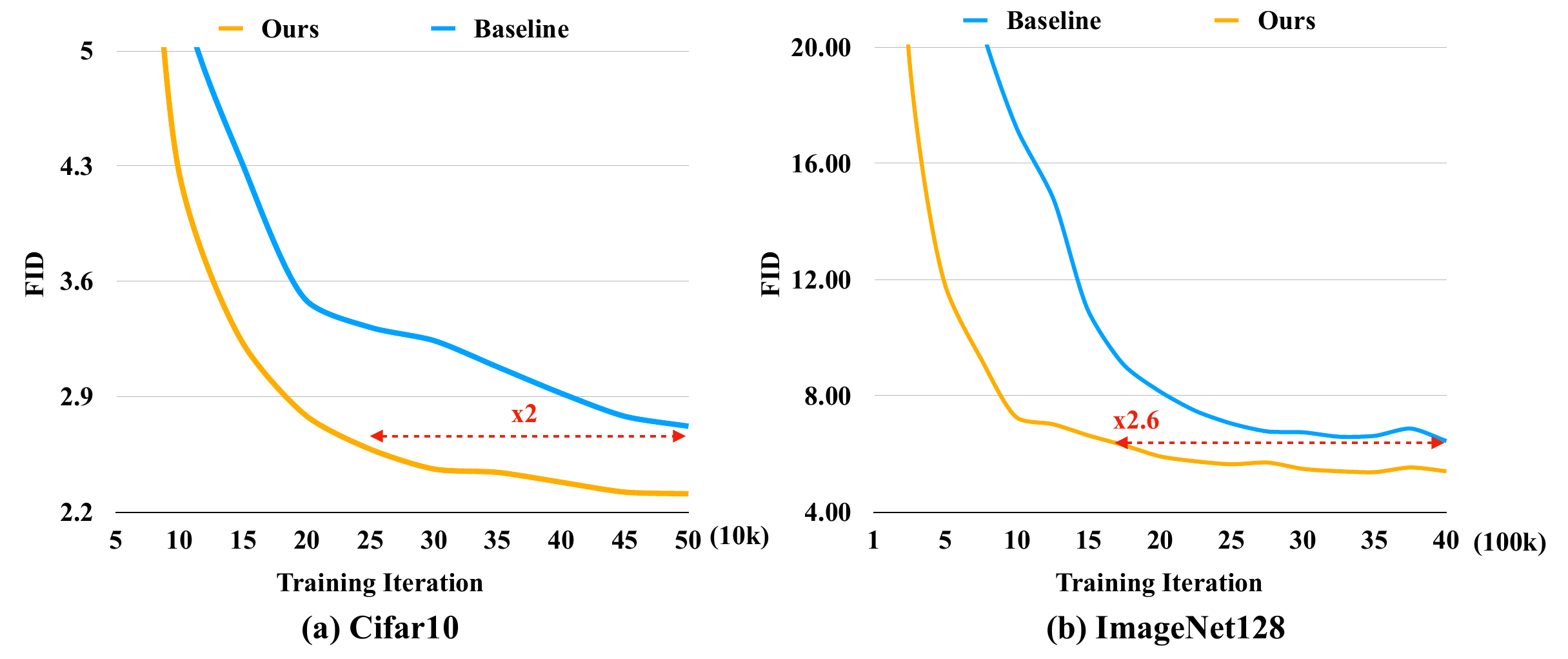}
  \caption{Illustration of the application of our optimization approach on different DMs. (a) Improved Diffusion \cite{nichol2021improved} trained on Cifar10 \cite{krizhevsky2009cifar}, (b) Guided Diffusion \cite{dhariwal2021beatgans} trained on ImageNet128 \cite{deng2009imagenet}. With our methods, these DMs achieve 2$\times$ and 2.6$\times$ speedup in training, respectively.}
  \label{fig:sota}
\end{figure*}

\begin{figure*}[thb]
  \centering
  \includegraphics[width=1.0\linewidth]{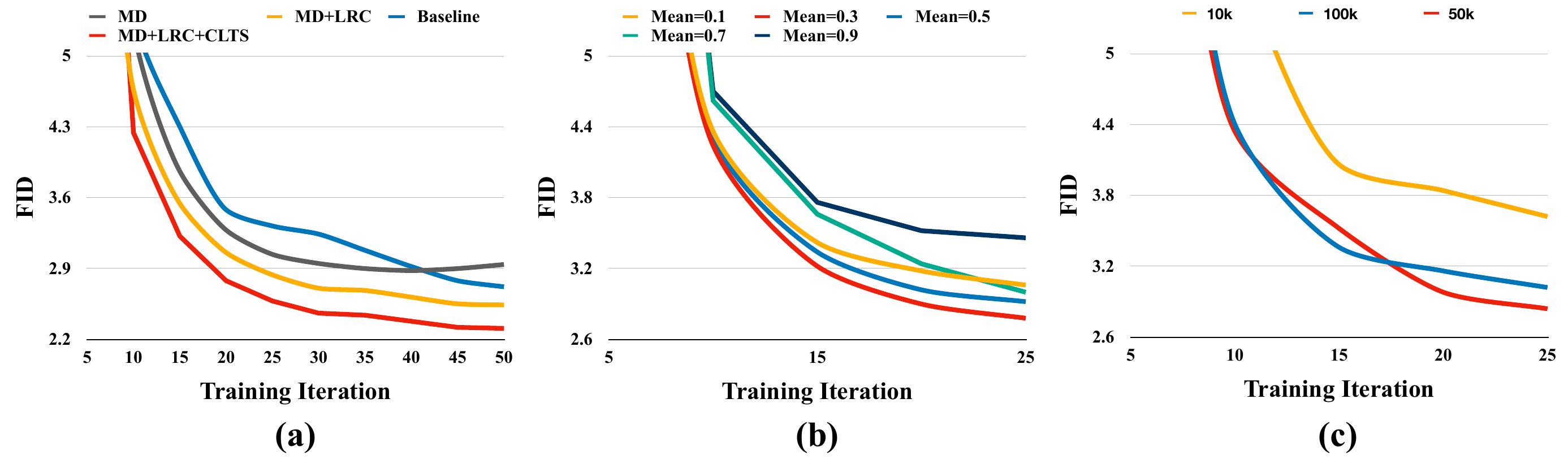}
  \caption{Ablation study, every model is trained on Cifar10. (a) shows the contributions of each module in our proposed methods. We compare the baseline model with momentum decay (MD), MD with learning rate compensation (LRC), and MDLRC with curriculum learning based timestep schedule (CLTS). (b) illustrates the influence of different mean $\mu$ in our proposed CLTS (Eq.~\ref{eq:gaussian}). (c) reflects the influence of values of different target iterations that we used in CLTS (Eq.~\ref{eq:gamma}).}
  \label{fig:ablation}
\end{figure*}

\subsection{Optimization of Momentum Schedule}\label{subsec:md}

Guided Diffusion \cite{dhariwal2021beatgans} was a pioneering work on DMs, incorporating advanced design concepts of BigGAN \cite{Brock2018LargeSG_biggan} to improve the performance and fidelity of DMs. However, they also adopted the large momentum setting, \emph{e.g.,} $\beta_1 = 0.9$, which is sub-optimal for DMs, whose loss landscape is highly smoothed. Applying a large momentum value in DMs would not only affect convergence efficiency but also cause oscillations, \emph{e.g.,} in Fig.~\ref{fig:sota}, the baseline model suffers oscillation in steps from 2.5M to 4M.

To address this issue, we propose the momentum decay with learning rate compensation (MDLRC). Following \cite{Chen2019DecayingMH}, the formulation of momentum decay is
\begin{equation}
    \beta_t = \beta_0 \cdot \frac{1 - \tau}{(1-\beta_0)+\beta_0(1-\tau)},\ \tau = \frac{\text{current iteration}}{\text{total iteration}},
\end{equation}
where $\beta_0$ is the initial momentum factor. Note that, because the total iteration is fixed, $\tau$ is not a hyper-parameter. However, since DMs usually use Exponential Moving Average (EMA) \cite{Oord2017NeuralDR_ema} to achieve better performance, simply applying momentum decay \cite{Chen2019DecayingMH} in DMs would over-amplification the weight of the current model in EMA, thus affecting the stability of the EMA model (we prove this in Appendix~\ref{app:momen_ema}).
Therefore, we compensate for the learning rate $l$:
\begin{equation}
    l_t = l_0 \cdot \frac{1 - \beta_0}{1 - \beta_t},
\end{equation}
where $l_0$ is the initiate learning rate, and $(1-\beta_0)$ and $(1-\beta_t)$ denote the factor of the initiate gradient and the factor of the current gradient, respectively. This compensation can ensure the weight of the current model in EMA remains consistent.

\section{Experiments}
In this section, we trained our optimized models on Cifar10 \cite{krizhevsky2009cifar} and ImageNet128 \cite{deng2009imagenet} datasets, following the hyper-parameter settings of Improved Diffusion \cite{nichol2021improved} and Guided Diffusion \cite{dhariwal2021beatgans}, respectively. The details of the hyper-parameter settings are as follows:

For Cifar10, we used a cosine timestep schedule, 4,000 timesteps, learning rate = 1e-4, and batch size = 128. We used an exponential moving average (EMA) rate of 0.9999 for all experiments. We implemented our models in PyTorch \cite{paszke2019pytorch}, and trained them on 8$\times$ NVIDIA 2080Ti GPUs, using 250 sampling processes. We used Adam optimizer, with $\beta_1=0.8, \beta_2=0.999$, which are based on the observation of the smooth landscape of diffusion models (DMs). The hyper-parameters of our proposed methods, namely curriculum learning based timestep schedule (CLTS) and momentum decay with learning rate compensation (MDLRC), are as follows: For CLTS, we set the mean value $\mu=1200$ (0.3 $\times$ total timesteps, the optimized mean value through ablation study), and the target iteration = $5\times10^4$. For MDLRC, we set the lower limit of $\beta_1 > 0.4$.

For ImageNet128, we used a linear timestep schedule, 1,000 timesteps, learning rate = 1e-4, and batch size = 256. We also used an EMA rate of 0.9999 for all experiments. We implemented our models in PyTorch \cite{paszke2019pytorch}, and trained them on 8$\times$ NVIDIA A100 GPUs, using 250 sampling processes. We used Adam optimizer, with initial $\beta_1=0.8, \beta_2=0.999$. The hyper-parameters of our proposed methods are as follows: For CLTS, we set the mean value $\mu=300$, and the target iteration = $3\times10^5$. For MDLRC, we set the lower limit of $\beta_1 > 0.4$.

\subsection{Ablations}

To evaluate the effectiveness of our proposed methods, we performed an ablation study. Fig.~\ref{fig:ablation} shows the results of the ablation study. Fig.~\ref{fig:ablation} (a) compares the performance of our proposed methods with and without each module, namely momentum decay (MD), learning rate compensation (LRC), and curriculum learning based timestep schedule (CLTS). The performance is measured by the FID score \cite{Heusel2017fid}, which is a widely used metric for assessing the quality and diversity of generated images. The results indicate that each module enhances the performance of the model, and the combination of all modules achieves the best FID score. Fig.~\ref{fig:ablation} (b) examines the effect of different mean values $\mu$ in our proposed CLTS (Eq.~\ref{eq:gaussian}). The mean value $\mu$ controls the most important timesteps in the Gaussian distribution. The results suggest that the optimal value of $\mu$ is around 0.3, which implies that the timesteps with the highest contribution to generation are not necessarily the most difficult ones to learn. Fig.~\ref{fig:ablation} (c) investigates the effect of different target iterations in our proposed CLTS (Eq.~\ref{eq:gamma}). The target iteration is a hyper-parameter that adjusts the speed of the Gaussian distribution emerging. The results demonstrate that the optimal value of the target iteration is around 100k, which means that the model needs about 100k iterations to fully adapt to the Gaussian distribution.

\begin{figure*}[thb]
  \centering
  \includegraphics[width=0.9\linewidth]{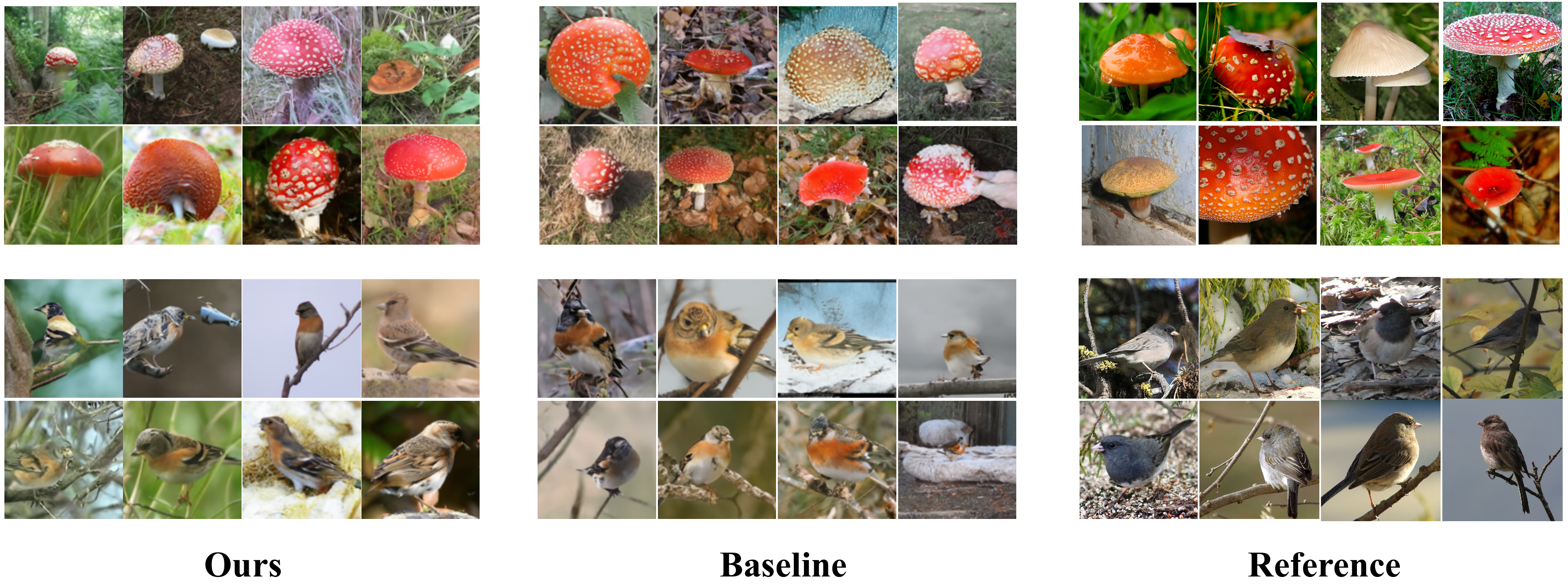}
  \caption{Comparisons of generated images. Both ours and the baseline are trained on Guided Diffusion \cite{dhariwal2021beatgans}, our model is trained in 1.6M iterations and the baseline is trained in 4M iterations. Reference is the original data of ImageNet \cite{deng2009imagenet}.}
  \label{fig:sota_images}
\end{figure*}

\begin{table*}[!t]
\centering
\caption{Comparing with state-of-the-art methods in ImageNet128 \cite{deng2009imagenet}, we use FID to evaluate the performance. Our method achieves the lowest FID score at each iteration.}
\label{tab:sota_imagenet}
\scalebox{0.9}{
\begin{tabular}{ccccc}
\hline
Methods & Iters=1M      & Iters=2M      & Iters=3M      & Iters=4M       \\ \hline
GD \cite{dhariwal2021beatgans}      & 17.18          & 8.14          & 6.63          & 6.04           \\
Min-SNR \cite{Hang2023EfficientDT}     &  13.53    &  6.49    &  6.11   &   5.81   \\
GD+Ours & \textbf{7.24} & \textbf{5.91} & \textbf{5.48} & \textbf{5.40} \\ \hline
\end{tabular}}
\end{table*}

\begin{table*}[!t]
\centering
\caption{Comparing with state-of-the-art methods in Cifar10 \cite{krizhevsky2009cifar}, we use FID to evaluate the performance. Our method achieves the lowest FID score at each iteration.}
\label{tab:sota_cifar}
\scalebox{0.9}{
\begin{tabular}{cccccc}
\hline
Methods & Iters=100k      & Iters=200k      & Iters=300k      & Iters=400k      & Iters=500k      \\ \hline
ID \cite{nichol2021improved}      & 5.40          & 3.48          & 3.05          & 2.72          & 2.60          \\
FDM \cite{Wu2023FastDM}     & 4.91          & 3.03          & 2.58          & 2.49          & 2.43          \\
ID+Ours & \textbf{4.24} & \textbf{2.81} & \textbf{2.46} & \textbf{2.38} & \textbf{2.31} \\ \hline
\end{tabular}}
\end{table*}

Based on the optimal settings, we trained our optimized models on Cifar10 \cite{krizhevsky2009cifar} and ImageNet 128 \cite{deng2009imagenet}, and compared them with the baseline models. Fig.~\ref{fig:sota} illustrates the results. The results reveal a significant acceleration of our optimized models, \emph{e.g.,} on Cifar10, our model achieves a 2$\times$ speedup compared with the baseline model, and on ImageNet128, our model achieves a 2.6$\times$ acceleration. Fig.~\ref{fig:sota_images} shows the visualization results of our methods, with 1.6M training iteration, our model shows competitive visual quality with the baseline models, which has trained 4M iteration. These results demonstrate the effectiveness and robustness of our proposed methods.

\subsection{Comparisons with state-of-the-art methods}
We compare our optimized models with state-of-the-art methods, Min-SNR \cite{Hang2023EfficientDT} and FDM \cite{Wu2023FastDM}. Table \ref{tab:sota_imagenet} compares the performance of our method with two state-of-the-art methods, Guided Diffusion (GD) \cite{dhariwal2021beatgans} and Min-SNR \cite{Hang2023EfficientDT}, on ImageNet128 \cite{deng2009imagenet}, and Table \ref{tab:sota_cifar} compares with Improved Diffusion (ID) \cite{dhariwal2021beatgans} and FDM \cite{Wu2023FastDM}, on Cifar10 \cite{deng2009imagenet}. The results demonstrate that our method achieves the lowest FID score at each iteration of both datasets, indicating that our method outperforms the existing methods in terms of image generation quality and speed.

\section{Conclusion}

In this paper, we have investigated the consistency phenomenon of diffusion models (DMs). We have attributed this phenomenon to two factors: the lower learning difficulty of DMs at higher noise rates, and the smoothness of the loss landscape of DMs. Based on this finding, we have proposed two strategies to accelerate the training of DMs: a curriculum learning based timestep schedule, and a momentum decay strategy. We have evaluated our proposed strategies on various models and datasets, and demonstrated that they can significantly reduce the training time and improve the quality of the generated images. Our work not only reveals the stability of DMs, but also provides practical guidance for training DMs more efficiently and effectively.


%% file: sec/X_suppl.tex
\clearpage
\setcounter{page}{1}
\maketitlesupplementary

\section{Preliminaries}\label{app:preliminaries}
We briefly review the definition of diffusion models from \cite{ho2020denoising}. The diffusion process is a method of transforming the original data $x_0$ into random noise by adding Gaussian noise $\epsilon$ at each timestep $t$, 
\begin{equation}\label{eq:forward_single}
    x_t = \sqrt{1 - \beta_t} x_{t-1} + \sqrt{\beta_t}\epsilon_t,
\end{equation}
where $x_t$ is the noisy data at timestep $t$, $\beta_t$ is the noise level at timestep $t$. The diffusion process starts from $t=0$ and ends at $t = T$, where $x_T$ is a pure noise vector. The noise level $\beta_t$ can be constant or vary across timesteps.

The reverse process is a way of recovering the original data from the noise by applying a denoising function $p(\cdot)$ at each timestep $t$,
\begin{equation}
    x_{t-1} \sim p(x_{t-1}|x_t, t).
\end{equation}
The reverse process starts from $t = T$ and ends at $t = 0$, $x_0$ is the generated data. The conditional distribution $p(x_{t-1}|x_t, t)$ is usually modeled by a neural network that outputs the mean and variance of a Gaussian distribution.

To train the neural network, we need to define a loss function that measures the discrepancy between the true distribution $q(x_{t-1}|x_t)$ and the learned distribution $p(x_{t-1}|x_t,t)$. One common choice is the KL divergence, 

\begin{align}
    L = &E_{x_0 \sim q(x_0), x_t \sim q(x_t|x_{t-1}), x_{t-1} \sim q(x_{t-1}|x_t)} \\
    &[\log q(x_{t-1}|x_t) - \log p(x_{t-1}|x_t, t)],
\end{align}
where $q(x_0)$ is the data distribution, $q(x_t|x_{t-1})$ is the forward diffusion distribution, and $q(x_{t-1}|x_t)$ is the reverse diffusion distribution. The expectation is taken over all possible pairs of $x_t$ and $x_{t-1}$ for each timestep $t$. While \cite{ho2020denoising} further simplified the above loss function and the new loss function can produce better samples in practice,
\begin{equation}\label{eq:simple_loss}
    L_{\text{simple}} = E_{x_0\sim q(x_0), \epsilon\sim \mathcal{N}(0, \textbf{I})}\left[||\epsilon - \epsilon_\theta(x_t, t)||^2 \right].
\end{equation}

To sample from the diffusion model, we need to sample a noise vector $x_T$ from a Gaussian distribution and then apply the reverse process to obtain $x_0$. This can be done by sampling $x_{t-1}$ from $p(x_{t-1}|x_t,t)$ at each timestep until $t=0$. The final sample $x_0$ is the output of the diffusion model.

\begin{figure*}[htb]
  \centering
  \includegraphics[width=0.9\linewidth]{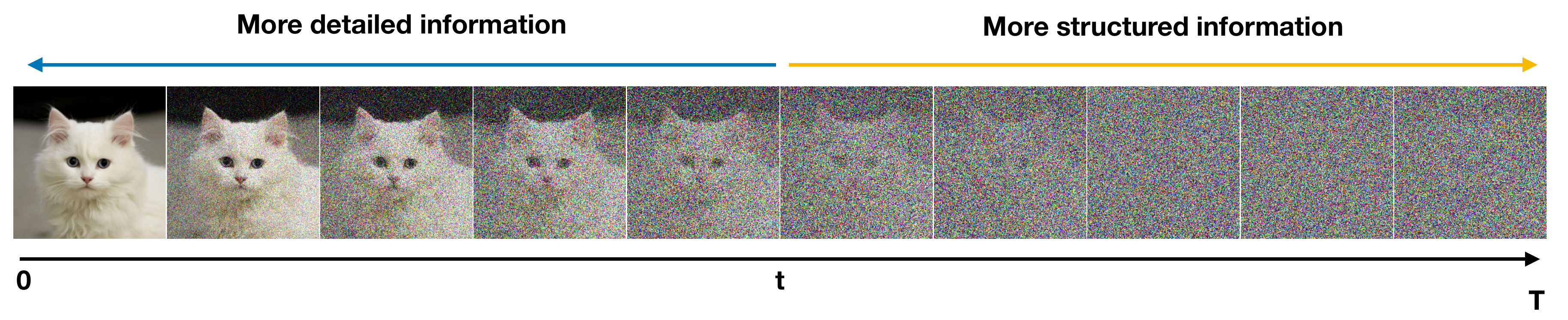}
  \caption{Visualization of the diffusion process.}
  \label{fig:d_process}

  \centering
  \includegraphics[width=0.9\linewidth]{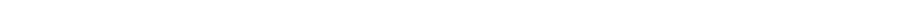}
  \includegraphics[width=0.9\linewidth]{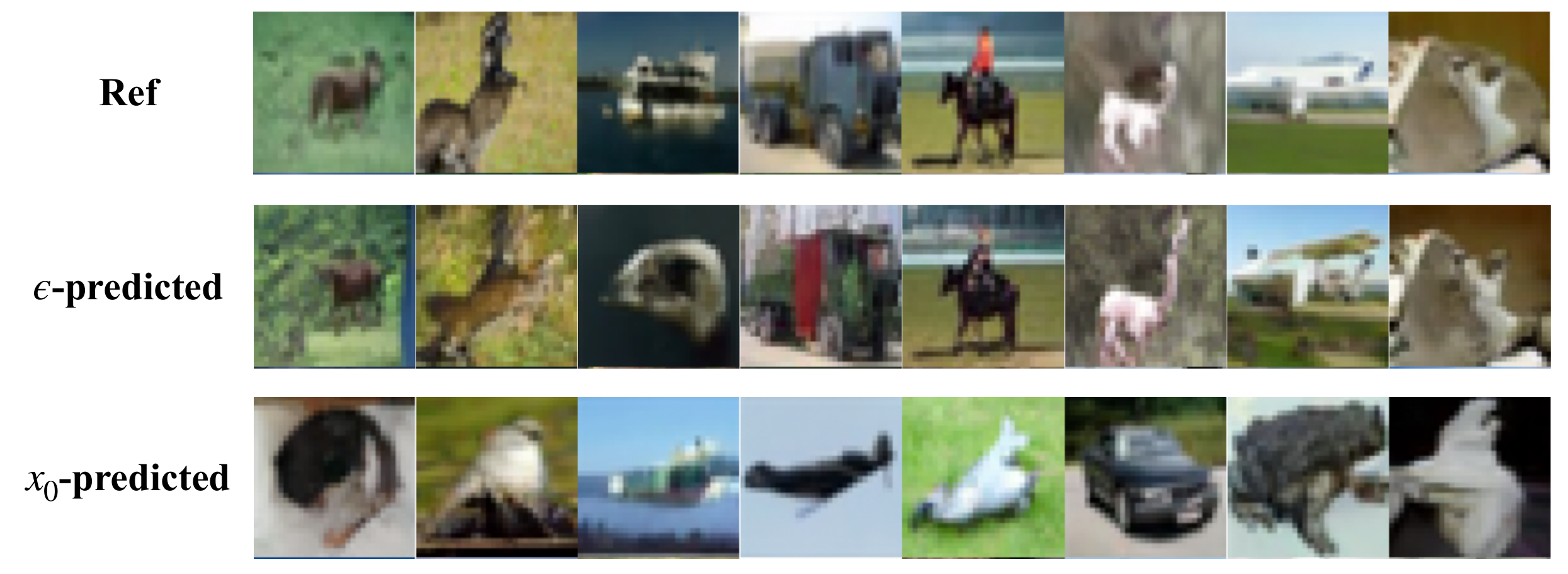}
  \caption{Illustration of consistency experiments of $\epsilon$-predicted and $x_0$-predicted DMs. The consistency phenomenon disappears when using $x_0$-predicted mechanism.}
  \label{fig:x0}
\end{figure*}

\newpage
\section{Relationship between Momentum Decay with Exponential Moving Average}\label{app:momen_ema}

The iterative formula of AdamW is
\begin{align}
    m_t &= \beta_1 m_{t-1} + (1-\beta_1) g_t \\
    v_t &= \beta_2 v_{t-1} + (1-\beta_2) g_t^2 \\
    \hat{m}_t &= m_t / (1-\beta_1^t) \\
    \hat{v}_t &= v_t / (1-\beta_2^t) \\
    \theta_{t+1} &= \theta_t - \gamma \hat{m}_t/\sqrt{\hat{v}_t}
\end{align}

The equation of EMA is
\begin{align}
    \hat{\theta}_{t+1} = \epsilon \hat{\theta}_t + (1 - \epsilon) \theta_t
\end{align}
As the decrease of $\beta_1$, the current gradient proportion increases, leading to the gradual dominance of the gradient in $m_t$, where $m_t$ is the main variable for updating $\theta_t$. Therefore, if the learning rate remains unchanged, the EMA model will be more influenced by the noise in the gradient.

\section{Comparisons of Loss Landscapes}\label{app:loss_land}
\begin{figure*}[h]
  \centering
  \includegraphics[width=0.9\linewidth]{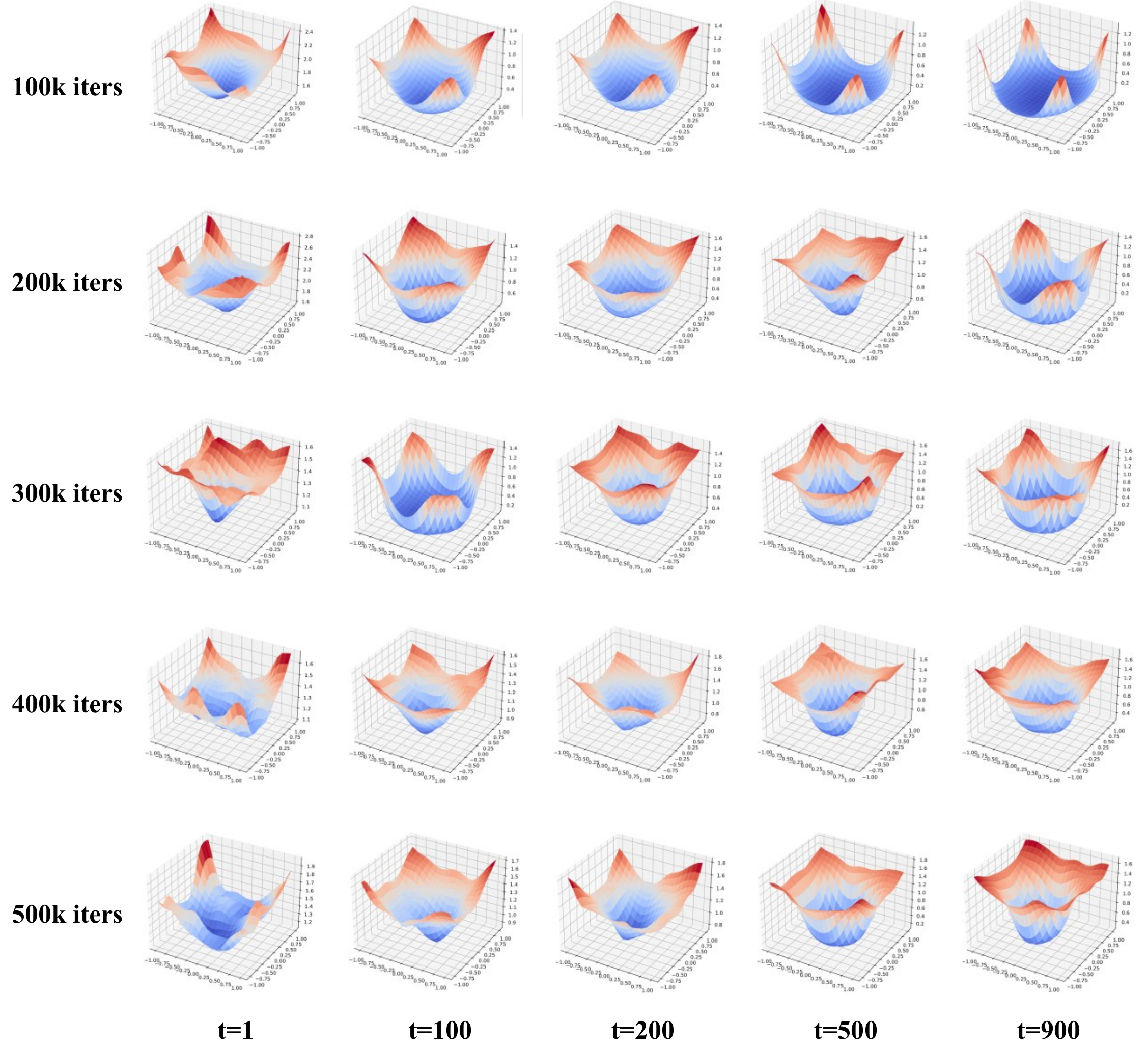}
  \caption{Loss landscape of diffusion models.}
  \label{fig:dm_land}
  \centering
  \includegraphics[width=0.65\linewidth]{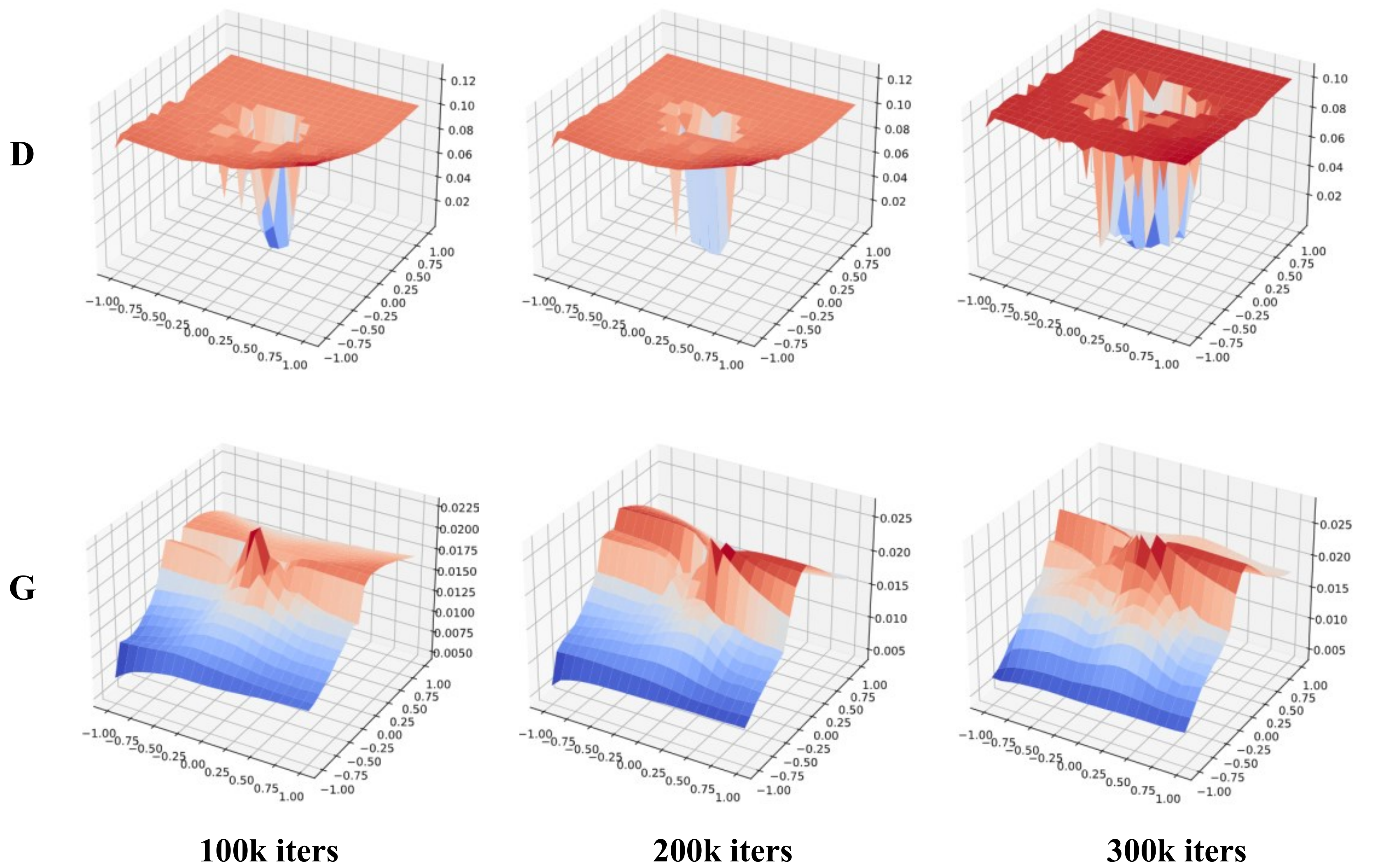}
  \caption{Loss landscape of generative adversarial networks.}
  \label{fig:gan_land}
\end{figure*}

In this section, we comprehensively illustrate the loss landscapes of DMs and GANs across various training iterations, as shown in Fig.~\ref{fig:dm_land} and Fig.~\ref{fig:gan_land}, respectively.

\section{Illustration of consistency experiments}\label{app:illus_consistency}
\begin{figure*}[h]
  \centering
  \includegraphics[width=0.8\linewidth]{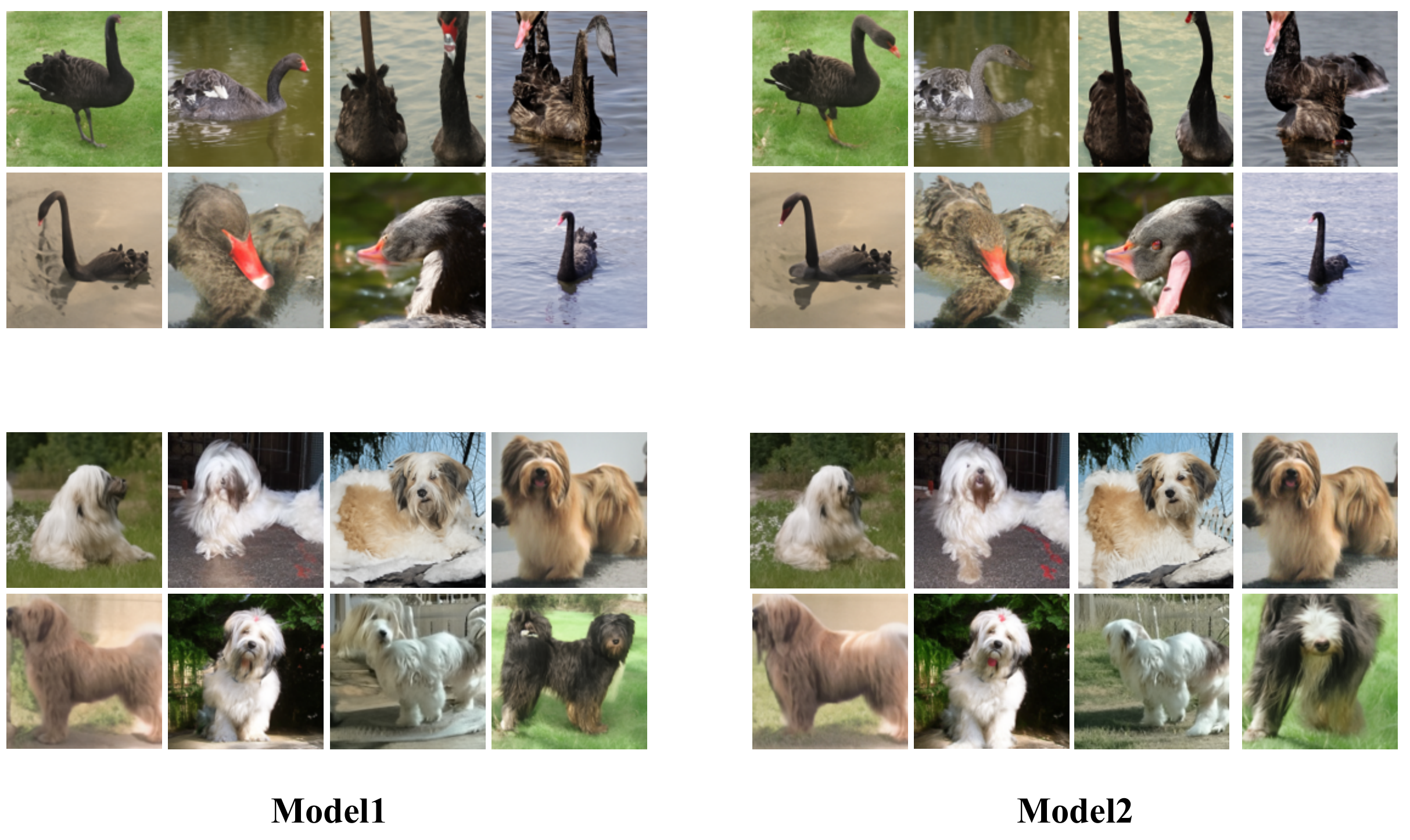}
  \caption{Consistency experiment results of Guided Diffusion at 128 resolution, each model trained on ImageNet. Obviously, DMs have a significant consistency phenomenon}
  \label{fig:imagenet_dm_consis}
\end{figure*}

\begin{figure*}[h]
  \centering
  \includegraphics[width=1.0\linewidth]{imgs/blank.pdf}
  \includegraphics[width=0.8\linewidth]{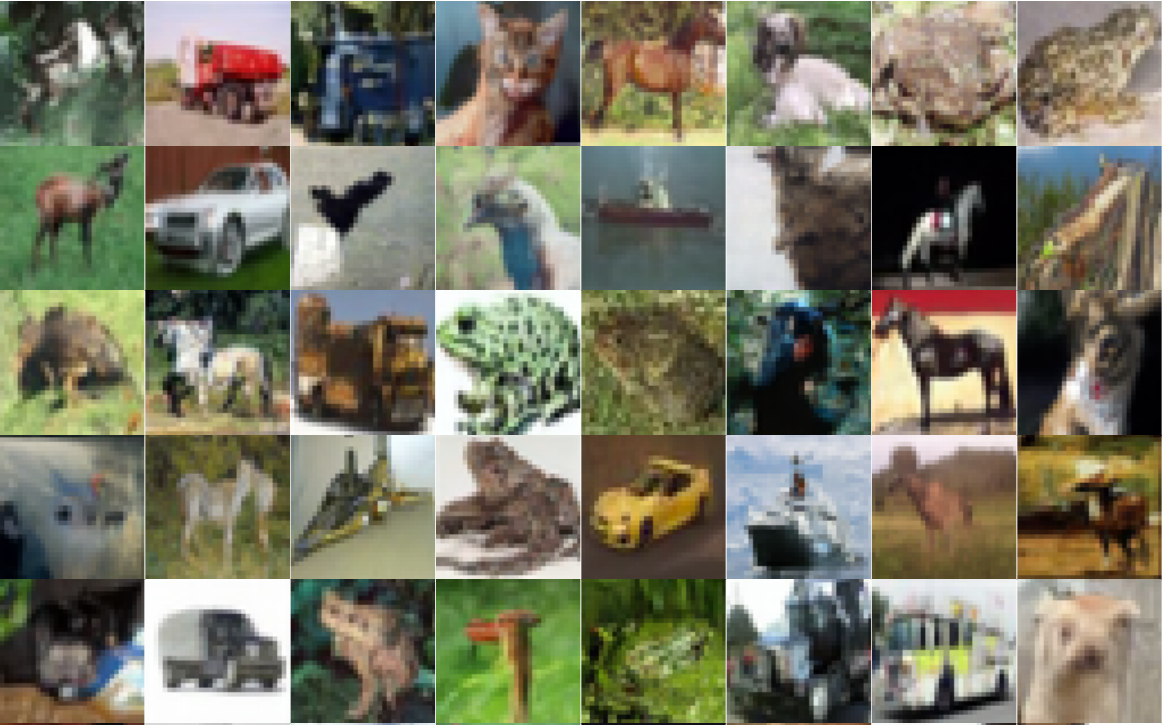}
  \includegraphics[width=0.8\linewidth]{imgs/blank.pdf}
  \includegraphics[width=0.8\linewidth]{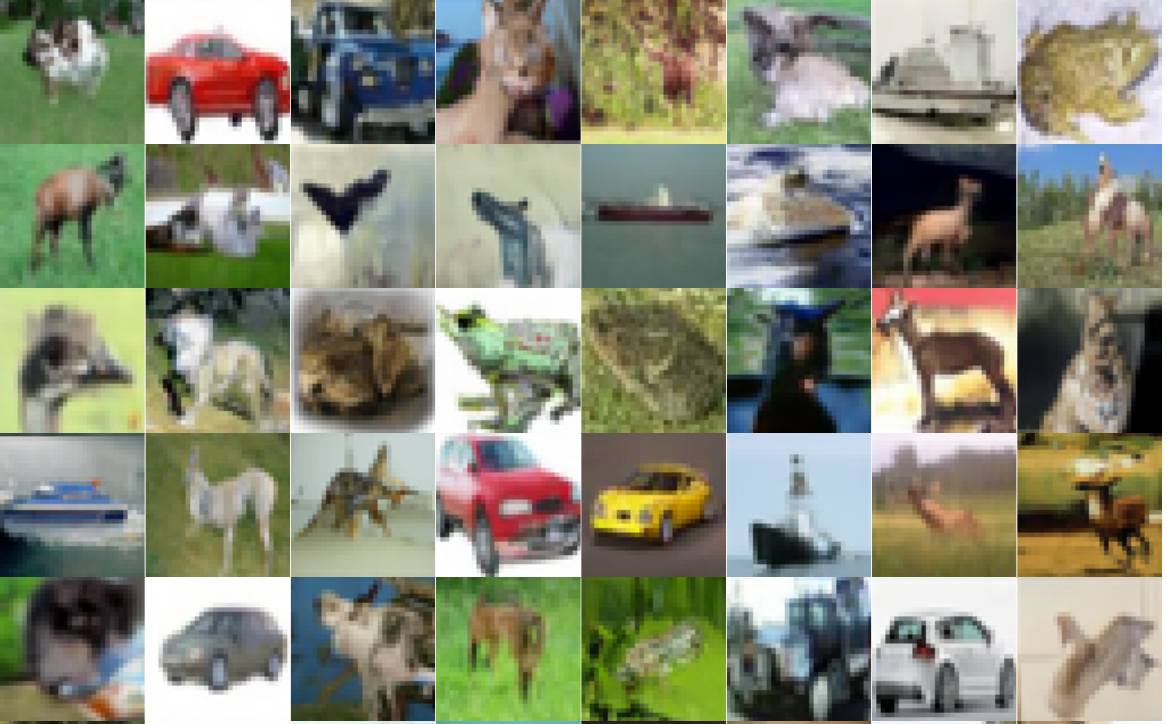}
  \caption{Consistency experiment results of Improved Diffusion at 32 resolution, each model trained on Cifar10.}
  \label{fig:cifar_dm_consis}
\end{figure*}

\begin{figure*}[h]
  \centering
  \includegraphics[width=0.8\linewidth]{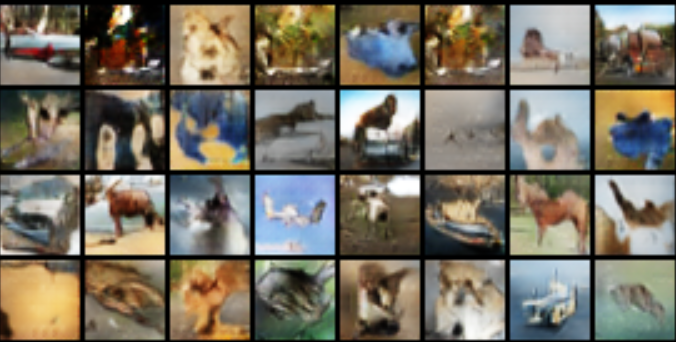}
  \includegraphics[width=0.8\linewidth]{imgs/blank.pdf}
  \includegraphics[width=0.8\linewidth]{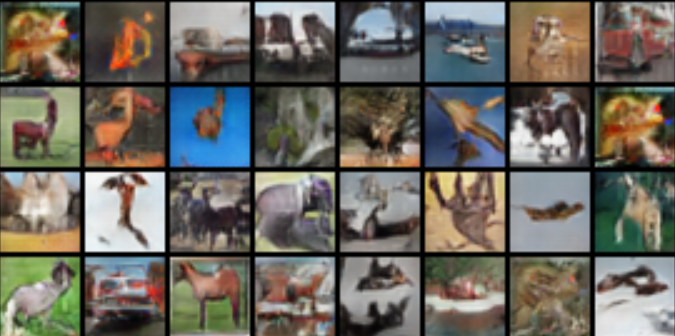}
  \includegraphics[width=0.8\linewidth]{imgs/blank.pdf}
  \includegraphics[width=0.8\linewidth]{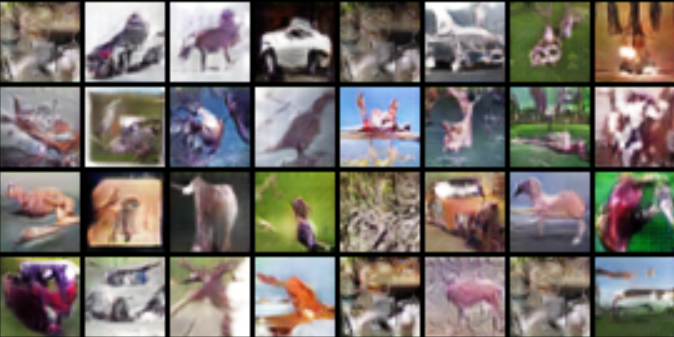}
  \caption{Consistency experiment results of DCGAN, three models with different initialization trained on Cifar10. Each model generates 32 images with the same sampling seed. Obviously, GANs have no consistency phenomenon.}
  \label{fig:cifar_gan_consis}
\end{figure*}

In this section, we provide more results of consistency experiments among DMs and GANs of different frameworks. The results of models trained on cifar10, can be seen in Fig.~\ref{fig:cifar_dm_consis}, Fig.~\ref{fig:imagenet_dm_consis}, and Fig.~\ref{fig:cifar_gan_consis}, from Improved Diffusion \cite{nichol2021improved}, Guided Diffusion \cite{dhariwal2021beatgans}, and DCGAN \cite{radford2015dcgan}, respectively.

\
\

\begin{figure*}[h]
  \centering
  \includegraphics[width=1.0\linewidth]{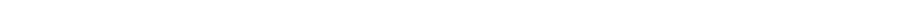}
  \includegraphics[width=1.0\linewidth]{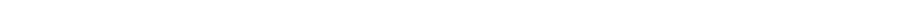}
  \includegraphics[width=1.0\linewidth]{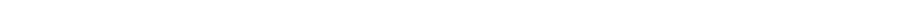}
  \includegraphics[width=1.0\linewidth]{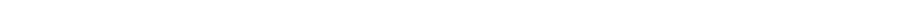}
  \includegraphics[width=1.0\linewidth]{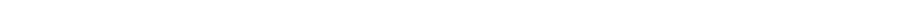}
  \includegraphics[width=1.0\linewidth]{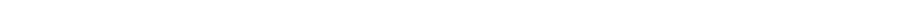}
  \includegraphics[width=1.0\linewidth]{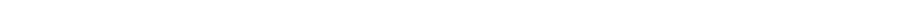}
  \includegraphics[width=1.0\linewidth]{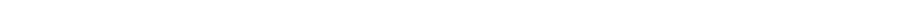}
  \includegraphics[width=1.0\linewidth]{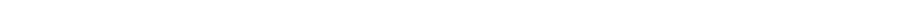}
  \includegraphics[width=1.0\linewidth]{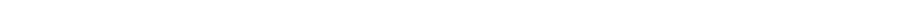}
  \includegraphics[width=1.0\linewidth]{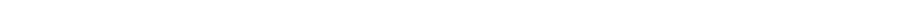}
  \includegraphics[width=1.0\linewidth]{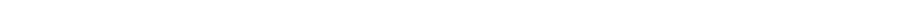}
  \includegraphics[width=1.0\linewidth]{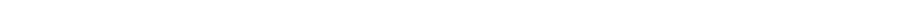}
  \includegraphics[width=1.0\linewidth]{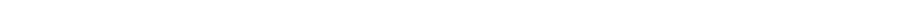}
  \includegraphics[width=1.0\linewidth]{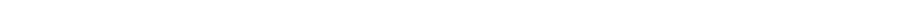}
  \includegraphics[width=1.0\linewidth]{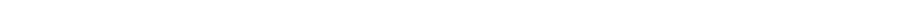}
\end{figure*}

\section{Illustration of Generated Results}\label{app:illustration}
\begin{figure*}[htb]
  \centering
  \includegraphics[width=\linewidth]{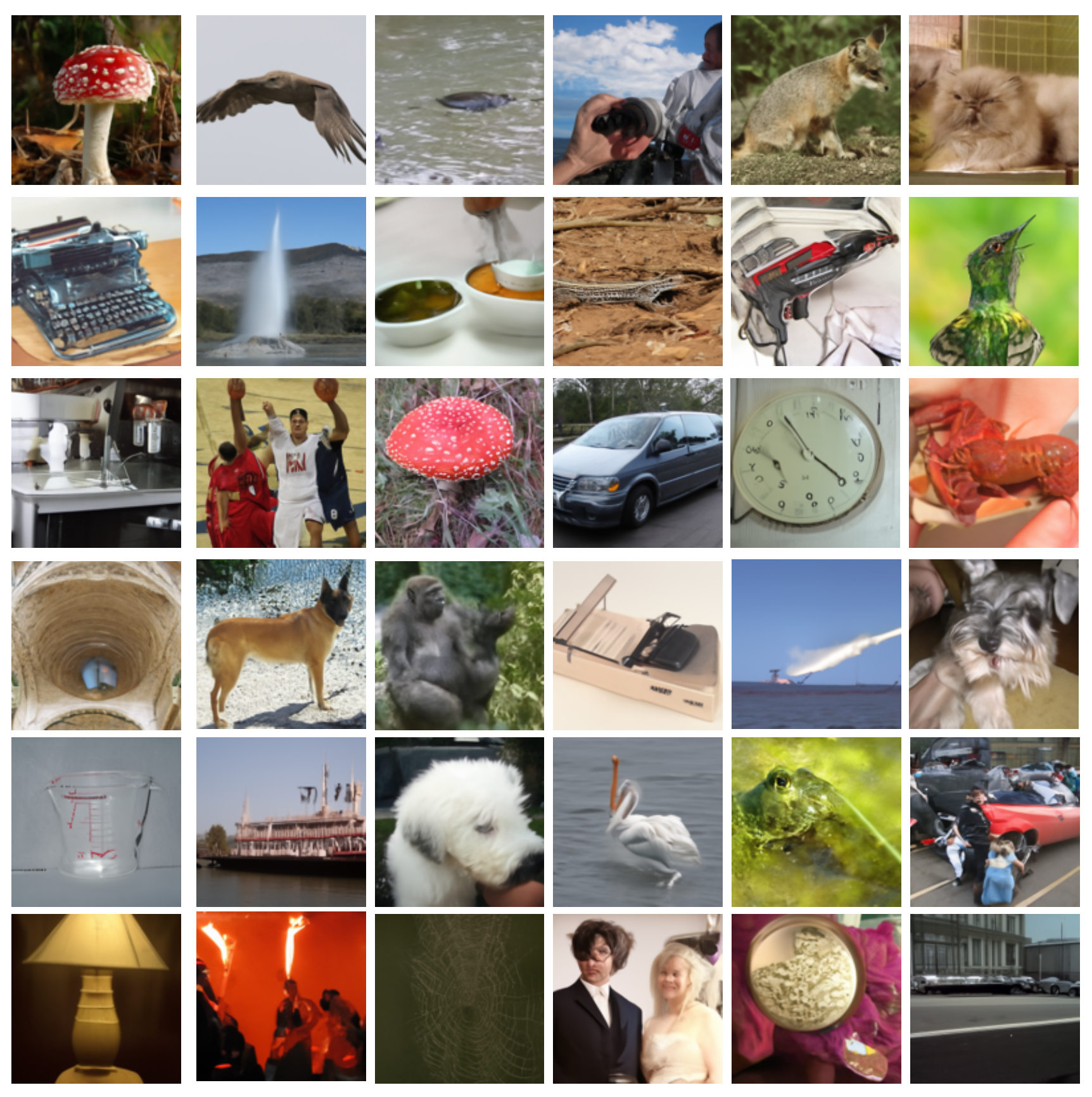}
  \caption{Results of our optimized diffusion model, which only trained 1.6M iterations on ImageNet128.}
  \label{fig:more_results}
\end{figure*}

We provide more generated results of our optimized DMs, as shown in Fig.~\ref{fig:more_results}. 

